\documentclass{article}
\usepackage{spconf,amsmath,graphicx}

\usepackage{enumitem}
\setlist{nosep, leftmargin=14pt}

\usepackage{mwe} 

\title{Tumor segmentation on whole slide images: training or prompting?}

\name{Huaqian Wu, Clara Brémond-Martin, Kévin Bouaou, Cédric Clouchoux}
\address{Witsee\\
33 Av. des Champs-Élysées, 75008 Paris, France}
\begin{document}
%
\maketitle
\begin{abstract}
Tumor segmentation stands as a pivotal task in cancer diagnosis. 
Given the immense dimensions of whole slide images (WSI) in histology, deep learning approaches for WSI classification mainly operate at patch-wise or superpixel-wise level.
However, these solutions often struggle to capture global WSI information and cannot directly generate the binary mask.
Downsampling the WSI and performing semantic segmentation is another possible approach. 
While this method offers computational efficiency, it necessitates a large amount of annotated data since resolution reduction may lead to information loss. 
Visual prompting is a novel paradigm that allows the model to perform new tasks by making subtle modifications to the input space, rather than adapting the model itself. 
Such approach has demonstrated promising results on many computer vision tasks. 
In this paper, we show the efficacy of visual prompting in the context of tumor segmentation for three distinct organs. 
In comparison to classical methods trained for this specific task, our findings reveal that, with appropriate prompt examples, visual prompting can achieve comparable or better performance without extensive fine-tuning.
\end{abstract}
\section{Introduction}
\label{sec:intro}

\begin{figure*}[ht]
  \includegraphics[width=\textwidth]{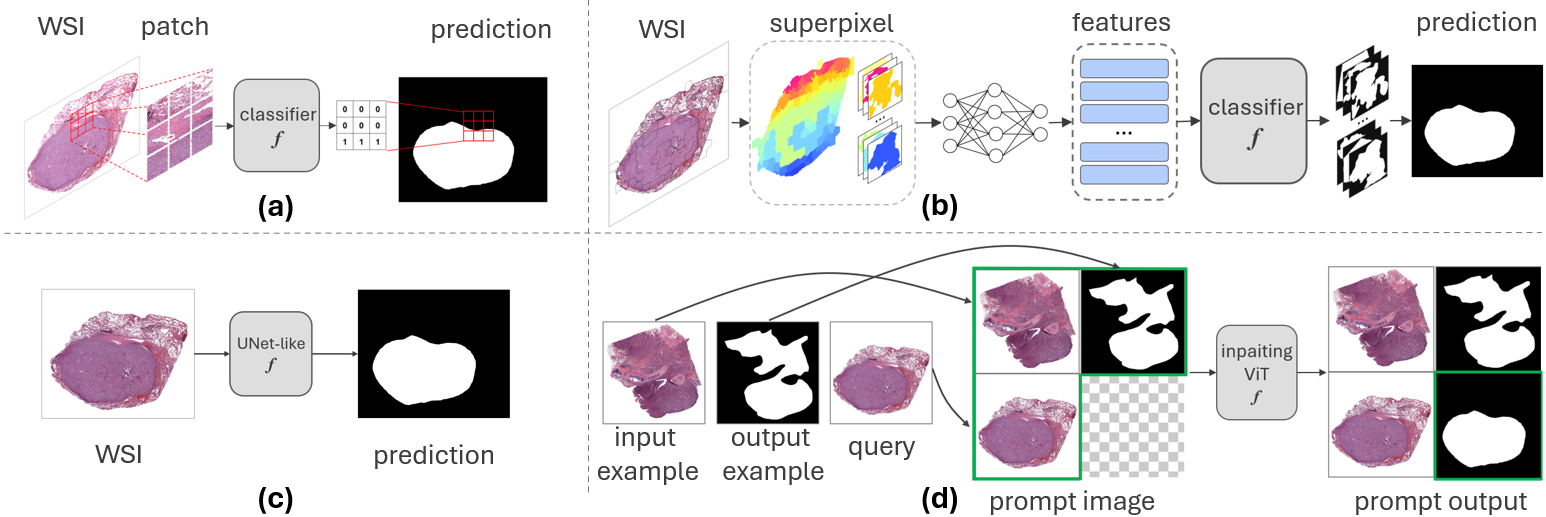}
  \caption{Illustration of different approaches. (a) patch-based classification. (b) superpixel-based classification. (c) semantic segmentation. (d) visual prompting.}
  \label{fig:methods}
\end{figure*}

The advent of whole-slide imaging has revolutionized pathology by enabling the digitization of entire glass slides of tissue samples, providing opportunities for efficient assessment and diagnosis. 
Cancer diagnosis involves the identification of tumor regions within whole slide images (WSIs). 
Tumor regions are often manually annotated by pathologists as regions of interest. This task is tedious and error-prone due to the large variability in tumor types and organs. 
In addition, tumor regions may have different shapes and sizes, and can also be mixed with other tissues, making the segmentation even more difficult.
Thus, there is a compelling need to develop Computer-aided diagnosis (CAD) systems that can automatically identify these regions. 

Tumor segmentation can often be considered as a binary classification task, where the input is a WSI and the desired output is a binary mask of tumor regions. 
However, gigapixel slide images present a challenge for CAD systems, since storing and processing such images is memory-intensive and computationally expensive. A feasible solution is to consider WSIs as a collection of smaller image segments and process them individually. For example, image patches and superpixels. Tumor segmentation can be conducted by classifying these image segments as tumor or non-tumor \cite{dos2021automated, bejnordi2015multi}.
Since image patches are small, the classification is computationally efficient. 
Superpixels are unions of pixels sharing similar color and texture characteristics.
Representing WSIs with superpixels reduces the number of individual pixels while preserving the structure information of the image. 
To conduct the classification, superpixel features are extracted and fed into a classifier \cite{bejnordi2015multi}. 
The choice of classifier can vary from a simple logistic regression to a deep learning model.
Deep learning models are often preferred in medical image analysis due to their effectiveness in capturing complex patterns and contexts.
In patch and superpixel classification, a subsequent reconstruction is necessary to generate the binary mask \cite{anklin2021learning,dos2021automated}. 

Both patches and superpixels represent image segments, potentially leading to misclassifications.
An alternative is to consider the WSI as a single entity and perform semantic segmentation. 
In contrast to tasks that demand high resolution (\textit{e.g.} nuclei segmentation), tumor segmentation primarily focuses on global characteristics and does not require such resolution. 
Hence, downsampling the WSI to a more manageable size becomes a practical strategy, mitigating computational demands and memory requirements. 
This approach offers a noteworthy advantage, as it yields a binary mask directly, avoiding the need of reconstruction step. 
In practice, semantic segmentation operates at the pixel level, employing a classical convolutional neural network (CNN) like U-Net. 
While such networks capture contextual information from neighboring pixels, their receptive field remains limited. 
For large tumor regions, a substantial receptive field becomes imperative to capture the full context information. 
Thanks to the self-attention mechanism that facilitates the capture of long-range dependencies in images, vision transformer (ViT) \cite{dosovitskiy2021image} has shown promising results in semantic segmentation as well as other computer vision tasks \cite{han2022survey}. 
However, ViT requires much larger datasets to outperform CNN \cite{dosovitskiy2021image}. 

In medical image analysis, one of the challenges for training deep learning models, especially ViT, resides in the few amount of labeled images available. 
To address this issue, a common strategy is to fine-tune ViT pre-trained on large natural image datasets. 
Nonetheless, training ViT is computationally expensive, and fine-tuning ViT for each downstream task significantly increases the training time. 

Recently, a new paradigm called visual prompting has emerged, bringing new perspectives for large pre-trained ViT \cite{jia2022visual,bahng2022exploring}. 
Prompting empowers the model to undertake new tasks by introducing subtle modifications to the input space, instead of adapting the model itself.
These modifications are known as prompts. Prompts can be learnable parameters, optimized during training \cite{jia2022visual,bahng2022exploring}. 
Prompts can also be in-context examples, serving as guidance for the model to produce desired output. 
This approach, called in-context learning, originates from GPT3 \cite{brown2020language} and formulates the machine translation as a text completion problem. 
Building upon this concept, \cite{bar2022visual} extended in-context learning to vision, by filling missing pixels in a grid-like image. The results demonstrated a strong generalization ability in object detection, colorization, and segmentation. 
SegGPT \cite{wang2023seggpt} is one of these solutions, dedicated to image segmentation and extensively trained on diverse datasets.

This paper aims to find a robust solution for tumor segmentation with the constraint of limited training data. 
Based on six WSIs from three organs, we conduct an evaluation of three training-based approaches (patch, superpixel classification, and semantic segmentation) and a visual prompting approach to perform tumor segmentation. 
Our findings reveal that visual prompting can obtain comparable performance to other approaches without fine-tuning the model.

\section{Materials and Methods}
\label{sec:method}

\subsection{Dataset}
\label{ssec:data}

We use the Cancer Genome Atlas (TCGA) dataset, with annotations provided by \cite{gao2020renal}. 
The training set contains six representative WSIs (average size: 75,378 $\times$ 98,914 pixels at 40$\times$), including two slides per tumor type (breast, lung, and kidney).  
The test set consists of three WSIs, representing three tumor types. 
Tumor segmentation aims to identify entire tumor regions. Therefore, all approaches, except patch classification, use the level 2  resolution of WSI, with a 16 downsampling factor. The training set is cropped into 256 $\times$ 256 pixels patches for patch classification and semantic segmentation.

\subsection{Patch-based classification}
\label{ssec:patch}

Figure \ref{fig:methods} illustrates different approaches for tumor segmentation.
Patch-based classification is executed at the patch level, yielding a prediction downsampled by the patch size. This approach requires a high resolution to keep a comparable prediction level as others methods. 
From six WSIs (average size: 18,844 $\times$ 24,728), a total of 35,834 patches are generated. The image patches are fed into EfficientNet-B5 \cite{tan2019efficientnet} pre-trained on ImageNet \cite{deng2009imagenet}. 
The binary mask is generated by reconstructing the classification results of the image patches.

\subsection{Superpixel-based classification}
\label{ssec:superpixel}

Superpixel feature extraction follows the pipeline in \cite{anklin2021learning}: we first generate superpixels using the Simple Linear Iterative Clustering algorithm \cite{achanta2010slic}, then extract features from the superpixels using MobileNet \cite{howard2017mobilenets} pre-trained on ImageNet \cite{deng2009imagenet}. 
In total, 2,460 superpixels are extracted from six WSIs. Each superpixel, representing a sub-region of the WSI, contains 10,240 features. The classifier is a simple CNN comprising three convolutional layers and two fully connected layers. 
The CNN takes the superpixel features as input. After the prediction, the binary mask is reconstructed by assigning the superpixel label to all corresponding pixels.

\subsection{Semantic segmentation}
\label{ssec:semantic}

Semantic segmentation is based on U-Net \cite{falk2019u}, with the encoder being replaced by ResNet34 \cite{he2016deep} pre-trained on ImageNet \cite{deng2009imagenet}. The six down-sampled WSIs are cropped into 2,531 patches to train the model. 

\subsection{Visual prompting}
\label{ssec:prompt}

We employ SegGPT \cite{wang2023seggpt} for visual prompting, as it is specifically designed for image segmentation. 
We choose a lung tumor image from train data as a prompt example. 
The test set is first normalized \cite{vahadane2016structure} to match the color distribution of the prompt image. 
Subsequently, test image is concatenated with the prompt image and fed into SegGPT to fill in the missing pixels, which correspond to the tumor regions.

The input and training objectives differ from one method to another and so does the training strategy. Multiple architectures have been tested, and only the best-performing ones are reported. 

\section{Results}
\label{sec:results}

\textbf{Implementation details.} All approaches are implemented in PyTorch on a NVIDIA Quadro P5000 GPU. 
Training is based on binary cross-entropy loss, Adam optimizer with a learning rate of 0.001. 
Training data are randomly divided into a training set (80\%) and a validation set (20\%). 
To enhance the robustness of the model, random augmentations such as rotation, horizontal and vertical flips, brightness adjustments, and elastic deformation are applied accordingly. 
Training is stopped when the validation loss does not decrease for 15 epochs. 
The implementation of visual prompting is based on the official code of SegGPT\footnote{https://github.com/baaivision/Painter/tree/main/SegGPT}. Dice score is used to evaluate the segmentation quality. 

\begin{table}[ht]
  \centering
  \caption{Dice score of different methods. The best results are in bold, the second best results are underlined.}
  \begin{tabular}{lcccc}
  \hline
  \textbf{image} & \textbf{patch} & \textbf{superpixel} & \textbf{semantic seg} & \textbf{prompt} \\
  \hline
  lung & 0.585 & 0.773 & \underline{0.821} & \textbf{0.920} \\
  breast & 0.329 & \textbf{0.821} & \underline{0.795} & 0.715 \\
  kidney & 0.912 & \textbf{0.940} & 0.825 & \underline{0.933} \\
  \hline
  mean & 0.609 & \underline{0.845} & 0.814 & \textbf{0.856} \\
  \hline
  \end{tabular}
  
  \label{tab:dice}
\end{table}

\textbf{Methods comparison.} Table \ref{tab:dice} reveals that patch classification yields the lowest Dice score across all images, averaging 0.609.
This score is 24.7\% lower than that of visual prompting. 
Visual prompting and superpixel classification stand out in the results, with the prompting method achieving a mean Dice score of 0.856, outperforming other methods. 
Prompting is also the best and the second-best method for lung and kidney images, respectively. 
Superpixel classification follows closely with a mean Dice score of 0.84 and stands as the top-performing method for breast and kidney images.
Even if semantic segmentation does not deliver the highest score, it demonstrates good robustness across different images, with only a 0.03 gap between its best and worst performances.

\begin{figure*}[ht]
  \centering
  \includegraphics[width=0.95\textwidth]{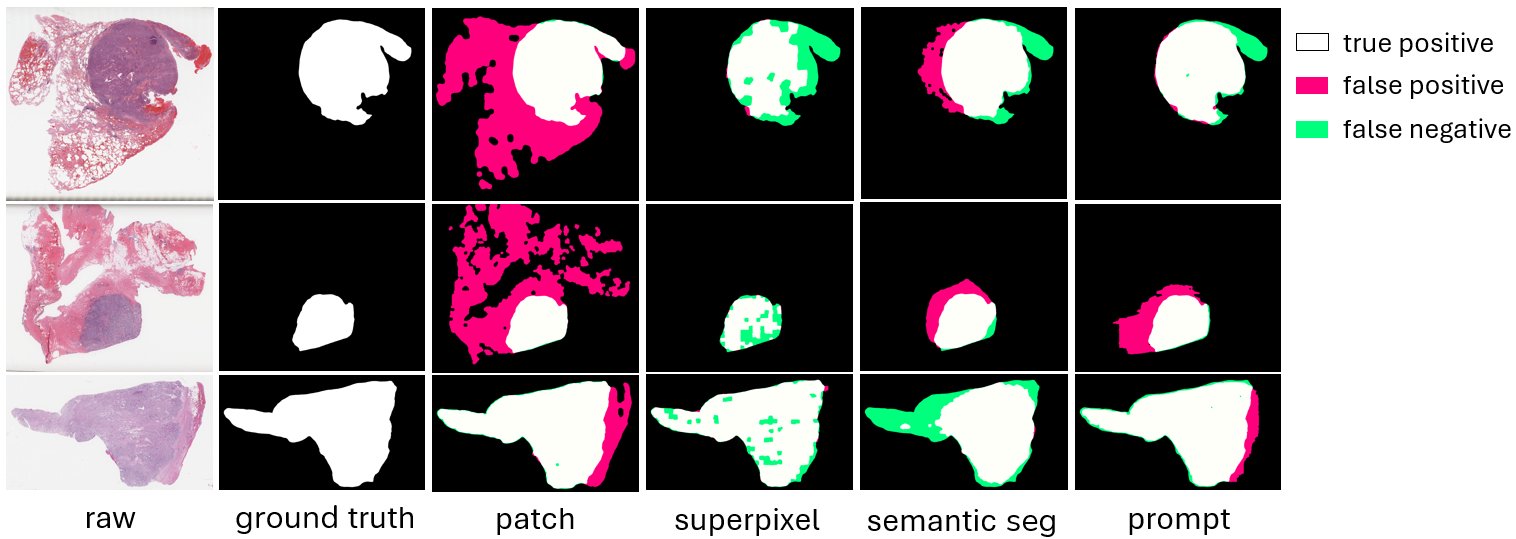}
  \caption{Visual comparison of different segmentation methods for three histological images. From top to bottom: lung, breast, kidney.}
  \label{fig:comparison}
\end{figure*}

Figure \ref{fig:comparison} provides a visual comparison of results. The suboptimal Dice score of patch classification can be attributed to tumor region over-segmentation. Superpixel classification, in contrast, suffers from under-segmentation. Semantic segmentation successfully identifies tumor regions, but it fails to determine precisely the contour. Both false positive and false negative pixels are observed around the tumor regions. The tumor regions in lung and kidney images are well captured by visual prompting, only a few false positive and false negative pixels are observed. However, it tends to over-segment the tumor regions in the breast image, resulting in a lower Dice score compared to superpixel classification and semantic segmentation. 

\begin{table}[h]
  \centering
  \caption{Processing time of different methods. The best results are in bold, the second best results are underlined.}
  \begin{tabular*}{\linewidth}{@{\extracolsep{\fill}} lcccc}
  \hline
  \textbf{method} & \textbf{patch} & \textbf{superpixel} & \textbf{semantic seg} & \textbf{prompt} \\
  \hline
  time (s) $\downarrow$ & 1153 & 166 & \textbf{1} & \underline{4} \\
  \hline
  \end{tabular*}
  \label{tab:time}
\end{table}

Table \ref{tab:time} compares the inference time required for a WSI. 
Among these methods, patch classification is the most time-consuming, with 19.22 minutes to process a single WSI. Superpixel classification proves to be six times faster than patch classification, requiring 2.77 minutes per WSI. 
Semantic segmentation is the fastest method, processing a WSI in only 1.02 seconds. 
Meanwhile, visual prompting is the second fastest, with a processing time of 4.11 seconds per WSI. 

\section{Discussion}
\label{sec:discussion}

This paper compares the performance of different approaches for small dataset tumor segmentation. 
The assessment is based on a training set (six WSIs) and a validation set (three WSIs) from three organs: lung, breast, and kidney. 

Superpixel classification stands out as the only method that does not over-segment tumor regions. 
It is based on a lower image resolution, substantially reducing the number of superpixels. 
Additionally, the simple three-convolutional-layer CNN offers quick training and imposes minimal memory requirements. 
Nonetheless, the additional time required for superpixel generation and feature extraction makes it impractical for real-world applications. 

Our findings reveal that patch classification not only emerges as the least effective but also the slowest. 
To ensure the same prediction resolution, we employ high-resolution WSIs for this method. 
While individual patches are fast to classify, the large amount of patches to cover the entire image leads to a significant increase in classification time. 

Furthermore, when evaluating the classification-based methods as illustrated in Figure \ref{fig:comparison}, it becomes evident that they both suffer from noise. 
The independent classification of each element, without consideration for neighboring patches or superpixels may produce this noise. 
In contrast, semantic segmentation and visual prompting process the entire image in a single pass, allowing them to capture the context information and produce smoother results. 
However, noise still persists in the semantic segmentation results for the lung and kidney images, which can be attributed to the limited receptive field of U-Net \cite{falk2019u}. 
Benefiting from the self-attention mechanism of ViT \cite{dosovitskiy2021image}, visual prompting capture information not only about the neighboring pixels but also about the entire image, and thus generates the least noisy segmentation. 

Processing the entire image at once yields a binary mask as the output, accelerating significantly the processing time. 
While semantic segmentation may not reach the highest accuracy, it is distinguished as the fastest and most robust method, making it a good candidate when time is a constraint. 
Regarding visual prompting, it outperforms particularly in lung image segmentation, this is probably linked to the use of a lung tumor image as prompt example. 
An effective prompt is expected to serve as a representative example, which holds true for the lung image but not necessarily for breast and kidney images. 
This finding highlights the importance of prompt selection. 
The variance in inference time between semantic segmentation and prompting may be attributed to differences in model size \cite{falk2019u,wang2023seggpt}. Semantic segmentation relies on ResNet34 \cite{he2016deep} with only 21 million parameters, while the prompting utilizes ViT \cite{dosovitskiy2021image} with 1.2 billion parameters. 

It is worth noting that ViT \cite{dosovitskiy2021image} is renowned for its data-hungry nature, often posing challenges in medical domain due to limited annotated data. 
In this study, visual prompting outperforms other classical methods customized for tumor segmentation, even without fine-tuning or adaptation. 
Using various prompt examples, the same model could extend its applicability to other tasks. This suggests that visual prompting could unlock the potential of large pre-trained ViT models in medical image analysis.
Nevertheless, vision prompting relies on appropriate prompt examples to guide the model in producing the desired output. 
In contrast, training-based methods offer greater flexibility, becoming deployable as soon as training is complete.

\section{Conclusion}

The aim of this paper is to find a robust solution for tumor segmentation. Public available models in vision are often trained on thousands even millions of examples \cite{kirillov2023segment}. Medical domain, however, presents unique challenges with its limited annotated data. Under such constraint, our experiments are conducted based on six training WSIs. 

The results demonstrate the efficiency of visual prompting in terms of segmentation quality and processing speed, despite the utilization of a single image as example, compared to the six images employed for training-based methods. 
This finding underscores the substantial potential of visual prompting in medical image analysis, especially for tasks with only a few annotation available.

Nevertheless, certain limitations should be acknowledged. The small training set constrains the choice of networks, preventing training-based methods from being boosted by more complex models. 
In contrast, visual prompting can benefit from the large pre-trained ViT. 
Further work is required to compare the performance of different approaches with an expanded training set. 
Another limitation lies in the use of a single prompt example, which may not cover test data variability. 
Given the exemplar-dependent nature of the prompting method \cite{zhang2023makes}, we believe the performance may be further improved with a suitable prompting for each specific organ.

\section{Compliance with Ethical Standards}

This research study was conducted retrospectively using human subject data made available in open access by TCGA. Ethical approval was not required as confirmed by the license attached with the data.

\bibliographystyle{IEEEbib}
\bibliography{refs}

\end{document}